\documentclass[conference]{IEEEtran}
\IEEEoverridecommandlockouts
\usepackage{cite}
\usepackage{amsmath,amssymb,amsfonts}
\usepackage{algorithmic}
\usepackage{graphicx}
\usepackage{textcomp}
\usepackage{xcolor}
\usepackage{cite}
\usepackage{balance}
\usepackage{url}
\usepackage{booktabs}
\usepackage{multirow}
\usepackage{multicol}
\usepackage{subcaption}

\usepackage{etoolbox}
\makeatletter
\patchcmd{\@makecaption}
  {\scshape}
  {}
  {}
  {}
\makeatother

\usepackage{multirow}
\def\BibTeX{{\rm B\kern-.05em{\sc i\kern-.025em b}\kern-.08em
    T\kern-.1667em\lower.7ex\hbox{E}\kern-.125emX}}

\begin{document}
\bstctlcite{IEEEexample:BSTcontrol}
\newcommand{\sid}[1]{\color{red}{ #1 } \color{black}}
\newcommand{\outline}[1]{\color{gray}{ #1 } \color{black}}
\newcommand{\fs}{$\texttt{fp16}$}

% \title{Benchmarking inference for large language models\\
\title{From Words to Watts: Benchmarking the Energy Costs of Large Language Model Inference\\
\thanks{DISTRIBUTION STATEMENT A. Approved for public release. Distribution is unlimited. This material is based upon work supported by the Assistant Secretary of Defense for Research and Engineering under Air Force Contract No. FA8702-15-D-0001, and United States Air Force Research Laboratory Cooperative Agreement Number FA8750-19-2-1000. Any opinions, findings, conclusions or recommendations expressed in this material are those of the author(s) and do not necessarily reflect the views of the Assistant Secretary of Defense for Research and Engineering, or the United States Air Force. The U.S. Government is authorized to reproduce and distribute reprints for Government purposes notwithstanding any copyright notation herein.}
}

\author{Siddharth Samsi\IEEEauthorrefmark{1}\textsuperscript{\textsection}, 
Dan Zhao\IEEEauthorrefmark{2}, 
Joseph McDonald\IEEEauthorrefmark{1},
Baolin Li\IEEEauthorrefmark{3},
Adam Michaleas\IEEEauthorrefmark{1},\\
Michael Jones\IEEEauthorrefmark{1}, 
William Bergeron\IEEEauthorrefmark{1},
Jeremy Kepner\IEEEauthorrefmark{1},
Devesh Tiwari\IEEEauthorrefmark{3},
Vijay Gadepally\IEEEauthorrefmark{1}\\
\IEEEauthorrefmark{1} MIT, 
\IEEEauthorrefmark{2} NYU,
\IEEEauthorrefmark{3} Northeastern University
}

\maketitle

\begin{abstract}
Large language models (LLMs) have exploded in popularity due to their new generative capabilities that go far beyond prior state-of-the-art. These technologies are increasingly being leveraged in various domains such as law, finance, and medicine. However, these models carry significant computational challenges, especially the compute and energy costs required for inference. Inference energy costs already receive less attention than the energy costs of training LLMs---despite how often these large models are called on to conduct inference in reality (e.g., ChatGPT). As these state-of-the-art LLMs see increasing usage and deployment in various domains, a better understanding of their resource utilization is crucial for cost-savings, scaling performance, efficient hardware usage, and optimal inference strategies. 

In this paper, we describe experiments conducted to study the computational and energy utilization of inference with LLMs. We benchmark and conduct a preliminary analysis of the inference performance and inference energy costs of different sizes of LLaMA---a recent state-of-the-art LLM---developed by Meta AI on two generations of popular GPUs (NVIDIA V100 \& A100) and two datasets (Alpaca and GSM8K) to reflect the diverse set of tasks/benchmarks for LLMs in research and practice. We present the results of multi-node, multi-GPU inference using model sharding across up to 32 GPUs. To our knowledge, our work is the one of the first to study LLM inference performance from the perspective of computational and energy resources at this scale. %and costs on such a scale.
\end{abstract}

\begin{IEEEkeywords}
Large Language Models, Natural Language Processing, Inference, Green AI, LLM, NLP, Deep Learning, Distributed Computing, Energy, Sustainability
\end{IEEEkeywords}

\section{Introduction}
\label{sec:into}
Generative models (GenAI) are able to produce new content from synthesizing text, images, and audio from which it's trained on. While GenAI is not entirely new, the recent application and broad availability of this technology via tools such as Stable Diffusion~\cite{stable-diffusion}, OpenAI's ChatGPT, Google's Bard and integration into the Microsoft Bing search engine has captured the imagination of the world and led to a massive surge in interest in deploying these types of models across a variety of domains ranging such as education, government, engineering, law, finance and many more. 

The popularity of these models has also put a spotlight on many societal concerns stemming from their usage. From ethical concerns ranging from violations of copyright laws~\cite{foster2022generative,smith2023} to safety concerns arising from the fact that these models are capable of hallucinating or fabricating information, concerns about these models in the educational and medical domain~\cite{mesko2023imperative, Zohny79}, their carbon footprint, and many more. 

In this paper, we focus primarily on understanding the significant amount of resources---time, computation, and energy---required for using and deploying some of the large language models (LLM) like those that underlie ChatGPT, Bard, etc. Several prior works have estimated the compute and energy costs of training language models. Works like \cite{strubell-etal-2019-energy} discuss the carbon footprint of language models such as BERT, ELMo, and precursors to larger models such as GPT-3 and GPT-4 that power some of the popular AI chatbots today. Others have also looked to larger language models; for instance, the largest NVIDIA Megatron-LM model required 3,072 A100 GPUs~\cite{shoeybi2020megatronlm,narayanan2021efficient,megatron-git} for its training. While the complete details (time and resources used) of compute required for training GPT-3/4 are not available, several estimates for training\cite{sevilla2022compute,semianalysis} and inference are publicly available. As industry attempts to shore up competitive moats and restrict information regarding their underlying LLM technologies, these details can become less reliable and available.  Compounding this issue, estimates for inference are even less readily available
\cite{DESISLAVOV2023100857} despite their significant share of energy costs and their likely larger impact on the environment \cite{greenworldAI}---especially since model inference calls can occur more frequently than training/fine-tuning for real-world deployments and applications. 

We present the results of our inference experiments on LLaMA~\cite{touvron2023llama}: an open sourced pre-trained large language models by Meta AI. The LLaMA model is available in a number of sizes but, in most cases, its larger variants typically require multiple high-end GPUs for both training and inference (assuming no further compression/distillation). While our emphasis is on characterizing the compute performance and energy used for multi-node, multi-GPU inference, we also include results from single node instances using smaller variants of the model as a baseline comparison. We hope our work will help illustrate some of the compute performance and energy utilization characteristics of LLM inference. We also hope that our experiments, analysis, and data on real-world hardware will spur further analysis, benchmarking, and more open dissemination of the systematic performance characteristics for a wider range of large models---especially under different kinds of hardware, data, and optimization strategies.

%~\cite{ren2023water} found that training ChatGPT required \sid{a ton of water... somthing something}.  
\section{Overview of Large Language Models}
\label{sec:llm}

The landscape of large language models (LLMs) and large foundation models (LFMs) has seen explosive growth in both the speed of development as well as complexity of ever larger models. Over the past several years, competition has been fierce and the pace un-relenting as AI research groups across private companies and academic institutions have developed new models whose performance continues to improve on a wide suite of natural language benchmarks but still requires significant amounts of compute and energy. We provide a brief overview of LLMs and LFMs below along with details around the specific LLM we use for our analysis.

\begin{figure}[!htbp]
    \centering
    \includegraphics[width=.9\linewidth]{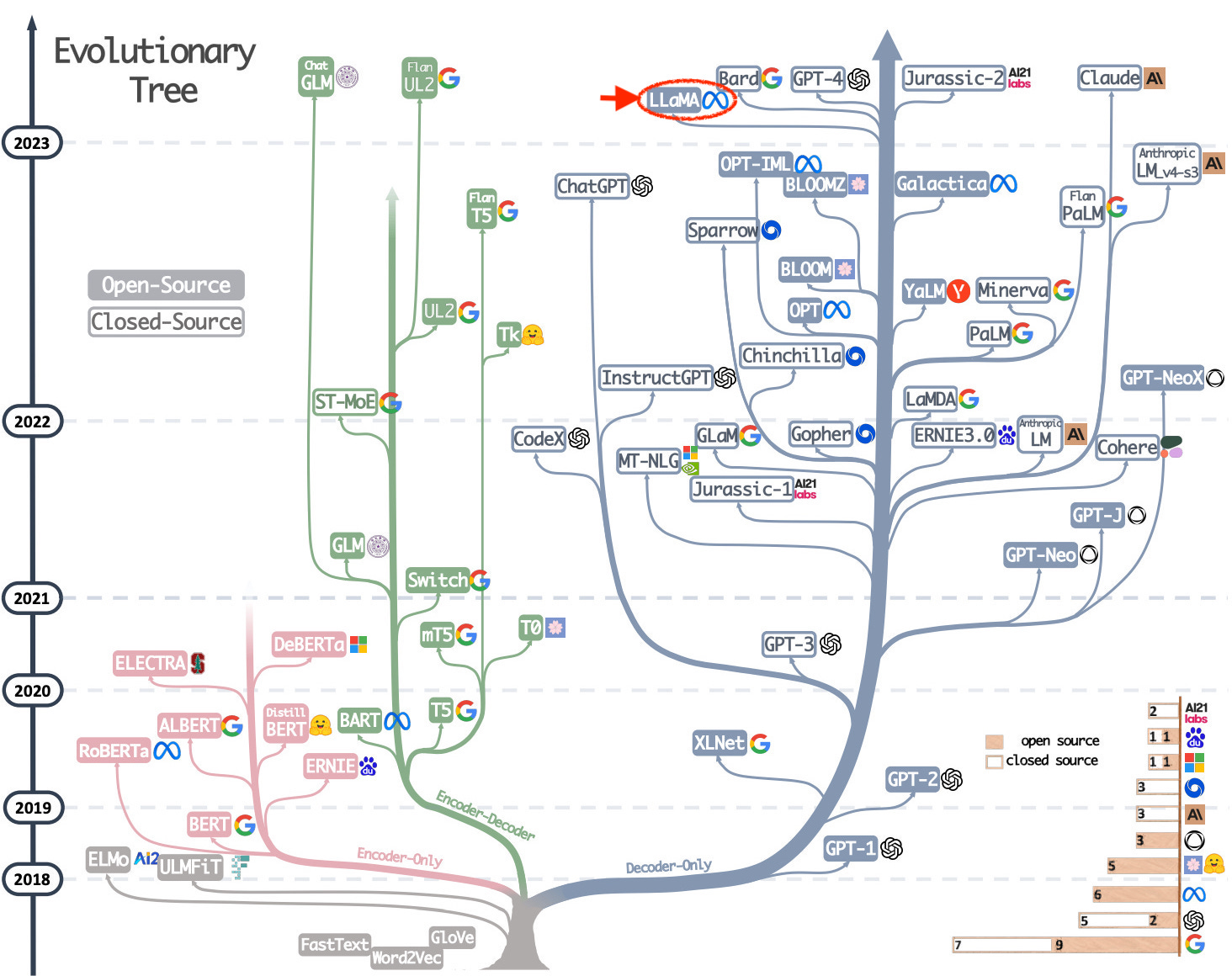}    
    \caption{\textbf{Development paths of LLMs:} A tree diagram illustrating the development of language models and foundation models from 2017 to early 2023. Pink branches indicate encoder-type language models, green indicates encoder-decoder hybrid models, and the dark grey indicates decoder-style models. The bar-plot on the bottom right tallies the number of open/closed source models developed by different companies/institutions.
    We study LLaMA (outlined by the red arrow and red circle in the diagram above) as an example of one of the more recent, modern, and state-of-the-art LLMs whose size/complexity resemble Google's Bard and OpenAI's GPT-4, all three of which were released around the same time (spring 2023). Original figure from \cite{llm_survey_pic}.}
    \label{fig:llm_survey}
\end{figure}

\subsection{Large Language Models \& Large Foundation Models}
 As seen in Fig. \ref{fig:llm_survey}, many different LLMs and foundation models exist---each with their own respective training setup, architectural modifications, purposes or use-cases, etc. Large language models and foundation models are best known for their sheer size, resource intensity (i.e., the amount of computational resources required for training/inference), and their impressive capabilities in tasks that include, but may not be limited to, natural language. 
 
 Typically, LLMs refer to language models containing on the order of hundreds of millions to billions of parameters that are trained on extremely large datasets of text. These models are also typically based on some variant of the original transformer architecture \cite{vaswani2017attention} usually leveraging the decoder half or a hybrid encoder-decoder architecture.
Large language models can be considered a subset of large foundation models; whereas LLMs focus almost exclusively on language data for their inputs and outputs, large foundation models include models that allow for multiple modalities such as image and text (e.g., GPT-4) or other modalities such as image generation (e.g., Stable Diffusion) or video generation (e.g., MidJourney). We refer to \cite{gozalobrizuela2023chatgpt} for a comprehensive review of the broad classes of GenAI and their capabilities. 

\subsection{LLaMA}
Developed by Meta AI and released in February of 2023, LLaMA \cite{touvron2023llama} (\textbf{L}arge \textbf{La}nguage \textbf{M}odel \textbf{M}eta \textbf{A}I) is a large language model (LLM) that relies on the traditional transformer architecture originally introduced in \cite{vaswani2017attention}. Most notably, the performance of LLaMA rivaled or exceeded that of GPT-3 on many NLP benchmarks and remains competitive with other state-of-the-art LLMs \cite{touvron2023llama}. Like other LLMs, LLaMA was pre-trained on a large collection of data including but not limited to CommonCrawl, Github, Wikipedia, etc.
 As of spring 2023, alongside other recently timed releases of state-of-the-art LLMs such as Google's Bard and OpenAI's GPT-4, LLaMA is competitive in its state-of-the-art performance across multiple tasks, making it an ideal workhorse for realistically studying and benchmarking inference.

LLaMA comes in four sizes characterized by the number of parameters: 7 billion (LLaMA 7B), 13 billion (LLaMA 13B), 33 billion (LLaMA 33B) and 65 (LLaMA 65B). LLaMA's model weights, across all of its variants, were publicly released under a non-commercial license, making it one of only a select few modern, state-of-the-art LLMs that have been publicly available.

To best understand the realities that lie behind the energy costs and throughput of state-of-the-art LLM inference, we focus our analysis on the largest available version of LLaMA---namely, LLaMA 65B. We also conduct analysis comparing the 7B and 13B LLaMA variants to establish the baseline performance of the smaller variants of the LLaMA model.

The largest model we focus our analysis on, LLaMA 65B, is a 65 billion parameter model with an effective model dimension of 8,192 and a total of 80 layers and 64 attention heads, trained over 1.4 trillion tokens. By focusing on the largest 65B version, we also hope to study inference at its fullest scale, controlling for and benchmarking phenomena that we may not observe on LLMs of smaller size or complexity. This way, we can realistically benchmark and study the dynamics, as well as the implications, of inference energy costs and through-put on a scale consistent with state-of-the-art LLMs that we see and use today. 
\section{Experimental Setup}
\label{sec:setup}
We conducted our experiments on the MIT Supercloud high-performance computing (HPC) system\cite{reuther2018interactive}. This heterogeneous HPC cluster consists of 448 compute nodes with dual Intel Xeon Gold 6248 CPUs with 384 GB of RAM and two NVIDIA Volta V100 GPUs with 32 GB of memory per node. Each node on the system has two independent back-end fabrics: a 100 Gb/s Intel Omnipath as well as a 25 Gb/s Ethernet interconnect using Mellanox ConnectX-4 adapters with all servers connected to a single, non-blocking Arista DCS-7516 Ethernet core switch. The GPUs, Omnipath, and Ethernet cards are all connected to PCIe slots that route directly to the Xeon processors without any intermediary PCIe switches. All experiments in this paper exclusively used the 25 Gb/s Ethernet interconnect. The system also includes 480 CPU-only nodes with Intel Xeon Platinum 8260 processors. In addition, four nodes with NVIDIA A100 GPUs were also available for experiments described in this paper. A summary of the hardware is shown in Table~\ref{tab:hardware}. All experiments described in this paper were run exclusively on NVIDIA GPUs.

\begin{table}[ht]
    \centering
    \caption{\textbf{Compute node configurations:} This table lists the types of hardware used for inference evaluations in our experiments. Each node consists of 2 CPUs and 2 GPUs in the configuration listed below. All GPUs are from NVIDIA.}
    \label{tab:hardware}
    \begin{tabular}{@{} cl*{7}c}%{p{.5cm} p{.5cm} p{.5cm} p{.15cm} p{.5cm} p{.5cm} p{.5cm}}%
    \toprule
     \multicolumn{3}{c}{CPU} & & \multicolumn{3}{c}{GPU}\\
     \midrule
     Type & Memory & TDP  & & Type & Memory & TDP  \\
          & (GB)   & (W)  & &      & (GB)  & (W) \\
    \cline{1-3} \cline{5-7}\\
     Intel Xeon \\ Gold 6248 & 384  & 150  & & V100 & 32 & 250 \\
     %& && & & & & \\
     %AMD \\ EPYC 7713 & 2,000 & 225 & & A100 & 40 & 250 \\
     & && & & & & \\
     Intel Xeon \\ Platinum 8358 & 503 & 240 & & A100 & 80 & 300 \\
     \bottomrule
    \end{tabular}
\end{table}

\subsection{Models}
Experiments were performed using open-source implementation of the pre-trained LLaMA 65B model available via request from Meta~\cite{touvron2023llama} and evaluation scripts available via GitHub~\cite{llama-git}. This implementation of the model uses Pytorch and the FairScale~\cite{FairScale2021} library to enable model sharding across multiple GPUs and nodes. For the models, we use a decoder temperature setting $\tau = 0.8$ and a top-$p$ value of 0.95 in attempts to align our settings with the general range of values that are typically used. In future work, we aim to study how varying decoding temperature, top-$p$, and other hyper-parameters may affect compute performance and energy usage during inference. While our main focus is on LLaMA 65B, we also examine LLaMA 7B and LLaMA 13B to characterize inference performance and energy under the bare minimum settings/resources required to run these models. 

\subsection{Datasets}
We used two datasets to evaluate inference performance. The first is an instruction following dataset used to fine-tune the Alpaca \cite{alpaca-git} model (from here on, this dataset is referred to as ``Alpaca'' in our paper which is not to be confused with the Alpaca model). This Alpaca dataset consists of 52,000 instruction-following tasks, instructions/questions where some have example inputs and some do not, that the model is asked to answer. The second dataset is GSM8K \cite{cobbe2021gsm8k}, consisting of 8,500 human crafted grade school math problems. The goal of using these two datasets is two-fold: (1) to evaluate the model on a diverse set of tasks in NLP and (2) evaluate how different types of data and its underlying dynamics can impact energy and inference performance. While natural language is more common in LLM usage and in LLM training data, increasingly new capabilities have been demonstrated in LLMs, including the ability to solve simple mathematical problems, provide/correct examples of code, and more. Math questions also differ considerably from questions posed in natural language which can result in smaller context windows, inputs/outputs of differing lengths, number of decoded tokens, etc. This, in turn, may impact inference performance in either throughput rates or energy costs. For this reason, our benchmarking experiments are conducted on both datasets.

For both datasets, we sample 4,096 inputs for our inference experiments. Using the entirety of the datasets would only serve to increase inference time and energy used for the experimentation unreasonably and did not provide any significant benefits to the study.% other than to crowd out other users on the HPC cluster. %in the interest of not excessively taking up resources on MIT Supercloud for too long so as to impact other users, as well as constraints on resource duration/availability, we randomly sample 4,096 observations from each dataset for our analysis.

\subsection{Evaluation}
\label{sec:eval}
Our goal is to evaluate the inference performance, latency, and inference energy costs of LLaMA 65B as a representative large language model that requires sharding across multiple GPUs. We intend this to be a preliminary analysis that will help guide more in-depth experiments and benchmarking for our future work. Our analysis also includes limited analysis of smaller LLaMA variants to illustrate inference performance and energy trade-offs in bare-minimum hardware settings: namely, LLaMA 7B and 13B. While we do not control for the correctness/quality of the outputs or the complexity of the inputs/outputs in studying trade-offs between inference energy and performance, we hope to account for this as an ablative study in future work. Similarly, we do not perform a comprehensive evaluation with different optimization techniques or inference settings available for LLMs such as modeling query arrival rates, model quantization, continuous batching, etc. which we also leave for future work.

Inference performance is measured in terms of rates: words, tokens, and responses per second or, equivalently, the number of words, tokens, and responses generated per second. When running inference with LLaMA, the model generates a string of text for each input until the length of the text hits a maximum generation length or a stop-word is encountered. The number of words are calculated by counting the number of words present in the output by splitting each output string on spaces. The number of tokens is calculated using LLaMA's own default tokenizer by counting the number of tokens in the tokenized output. Lastly, the number of responses per second or the response rate is calculated using the total number of responses and the total time needed to run inference over the input data. 

We monitor GPUs using the \texttt{nvidia-smi}~\cite{nvidia-smi} and \texttt{NVIDIA DCGM}~\cite{dcgm} utilities to study GPU utilization, energy, power draw, etc. during our experiments. The \texttt{nvidia-smi} utility is used to capture GPU usage over time at 100ms intervals and the \texttt{DCGM} monitoring tool is used to capture aggregate GPU energy in Joules for the rank-0 node. For a multi-node, multi-GPU model, we multiply the rank-0 energy by the number of nodes used. Maximum power draw on GPUs is capped at 250 Watts unless otherwise stated. Due to limits on resource availability, we mainly use V100 GPUs for larger-scale distributed experiments (i.e., for 8, 16, and 32 shards) and A100 GPUs for smaller scale experiments.

Inference energy metrics are calculated by combining the inference metrics above with the energy data collected from our GPUs using NVIDIA's utilities described above. Specifically, energy per second is defined as the total aggregate GPU energy spent from a single experiment/job (across all shards) divided by the total run time of that experiment/job in seconds. A single experiment/job denotes a single run through all 4,096 prompts under a specified batch size. Energy per token and energy per response are similarly defined as total energy divided by the number of decoded output tokens and the number of responses as defined above, respectively.

\section{Results}
\label{sec:results}

\subsection{Baselines: LLaMA 7B, 13B, \& 65B}
\subsubsection{Inference Performance}

We begin our analysis with a baseline comparison of LLaMA 65B with smaller-scale LLaMA models: LLaMA 7B and 13B. The goal is to understand the following: what do inference performance and energy trade-offs look like for the different sizes of LLaMA under the bare-minimum set of resources required to have them running inference? This question can be important for researchers and users who have may not have limitless computational resources and hardware acceleration or may be constrained in terms of GPU memory, etc. 

Given the sizes of the models, the size of the data, and the hardware memory limits, we only show results from experiments that were possible for a given combination of parameters (i.e., for some models, certain combinations of batch size and number of shards are infeasible due to memory limits of the underlying GPUs). Table~\ref{tab:limits} shows the bare minimum hardware requirements for each LLaMA variant and the maximum batch size possible for each combination, assuming no further model compression, optimization, quantization, distillation etc.

\begin{table}[ht]
    \centering
    \caption{\textbf{Baseline configurations for LLaMA 7B, 13B, and 65B:} This table lists the bare minimum hardware required for different models and the maximum batch size possible given the bare minimum hardware for a max response length of 256. These limits are imposed by a combination of GPU memory, model size, response length and the number of GPUs. While the 65B model can sharded across 6 V100 GPUs, we use 8 since the model architecture makes it better suited for balanced sharding across 8 GPUs.}
    \label{tab:limits}
    \begin{tabular}{cccccc}%{p{.5cm} p{.5cm} p{.5cm} p{.15cm} p{.5cm} p{.5cm} p{.5cm}}%
    \toprule
     Model Size & \multicolumn{2}{c}{V100 32GB} && \multicolumn{2}{c}{A100 80GB}\\
     \midrule
                & Count & Max. Batch size && Count & Max. Batch size \\ 
    \cline{2-3} \cline{5-6}\\
     7B  & 1 & 64 && 1 & 64 \\
     13B & 2 & 64 && 1 & 64 \\
     65B & 8 & 64 && 4 & 128\\
     \bottomrule
    \end{tabular}
\end{table}

\begin{figure}[h]
    \centering
    \begin{subfigure}{\linewidth}
        \includegraphics[width=.98\linewidth]{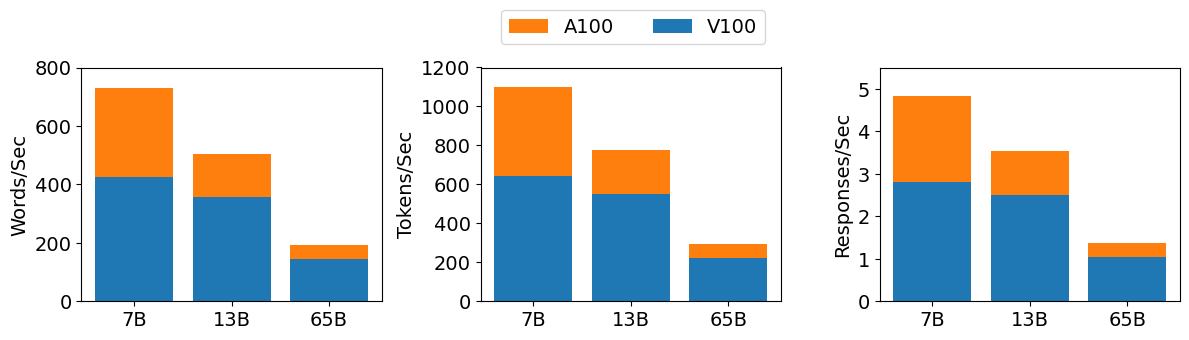}
        \caption{Results from the Alpaca dataset.}
        \label{subfig:alpaca-baseline}
    \end{subfigure} 
    \hfill
    \begin{subfigure}{\linewidth}
        \includegraphics[width=.98\linewidth]{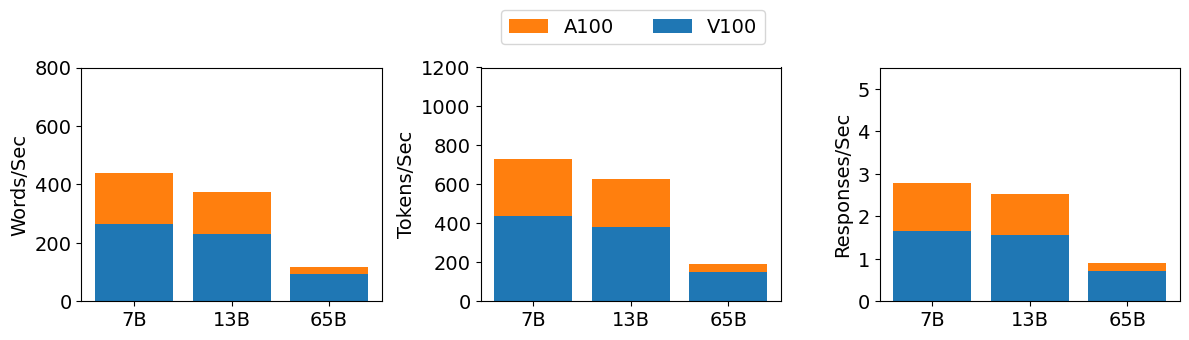}
        \caption{Results from GSM8K dataset}
        \label{subfig:gsm8k-baseline}
    \end{subfigure}
    \caption{\textbf{Baseline comparison of inference performance/latency between LLaMA 7B, 13B and 65B}: inference performance comparisons on the minimum set of hardware required to run inference (see Table \ref{tab:limits})  across model sizes and between V100s and A100s.}
    \label{fig:baselines}
\end{figure}

% now some plots comparing 7, 13, 65b model performance
With these limits in mind, we present the inference performance of LLaMA 7B, 13B, and 65B on the Alpaca and GSM8K datasets with the bare minimum hardware settings in Figure \ref{fig:baselines}. The plots in Figure \ref{fig:baselines} show a baseline comparison of inference performance of the three LLaMA variants on both the V100 and A100 GPUs respectively. For each model, in line with the spirit of the bare minimum settings, inference is done with a batch size of 64 and an maximum generation length of 256. The 7B model was run on a single GPU and 13B on two GPUs in each case whereas the 65B model was run on 8 V100 GPUs and 4 A100 GPUs respectively due to the size of the model and available memory on the GPU(s). 

As expected, we observe that the A100 outperforms V100 on both the Alpaca and GSM8K datasets: particularly for the smaller LLaMA 7B and 13B, we see anywhere from a 2 times (7B) to a 1.25 times increase (13B) in inference latency on the A100 when compared to the V100 across words per second, tokens per second, and responses per second. Faster response rates and inference are likely due to the fact that the number of computations, directly related to the number of parameters of said model, involved in the 7B and 13B models are significantly lower than the 65B model. We do note that for LLaMA 65B, we see a much smaller improvement in using the A100 over the V100; however, since the 65B model requires sharding across two (A100) or four (V100) compute nodes at the mininum, this could result in additional latency to each forward pass of the model, explaining the smaller improvements. We also observe that while LLaMA 7B exhibits a considerable improvement in inference throughput on both Alpaca and GSM8K with the A100, the improvement is much larger for Alpaca than GSM8K. This can also be attributed to the different complexities of inputs from each dataset. 

% subfigure help - https://latex-tutorial.com/subfigure-latex/

\subsubsection{Inference Energy}
\begin{figure}[hbtp!]
    \centering
    {{\includegraphics[width=7cm]{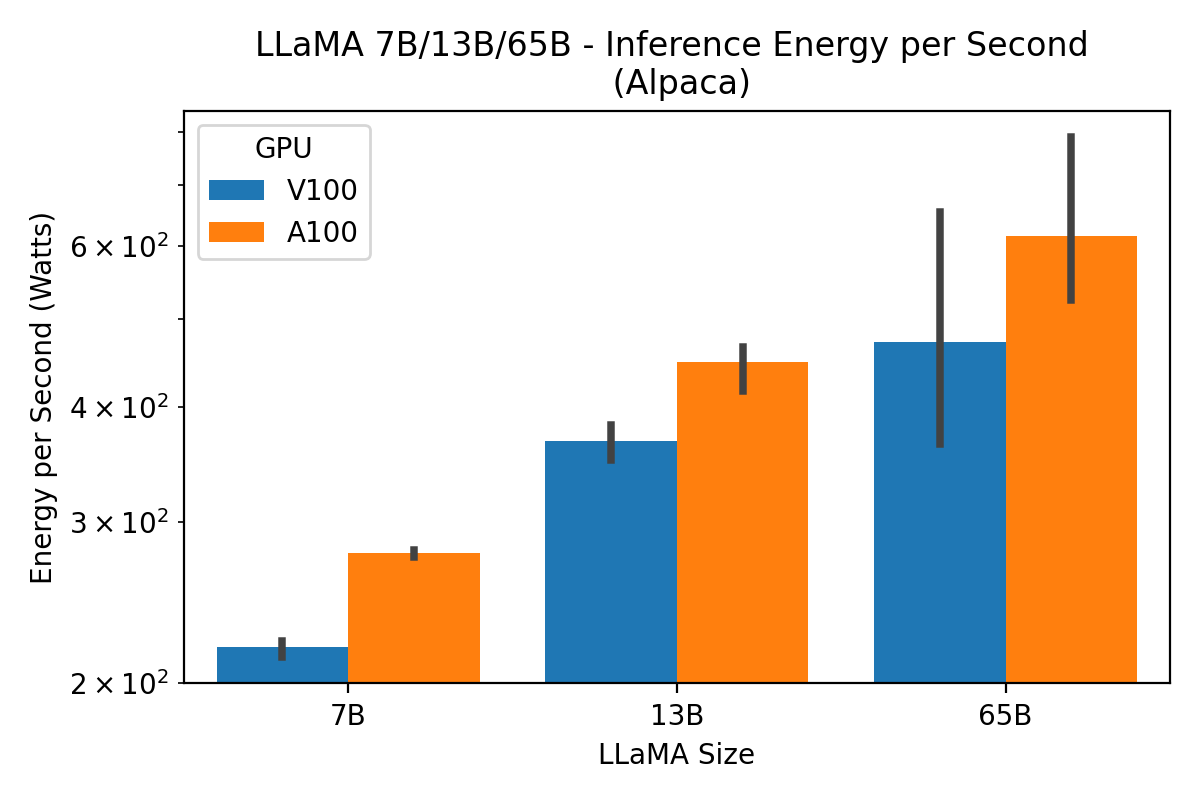} }}
    \qquad
    {{\includegraphics[width=7cm]{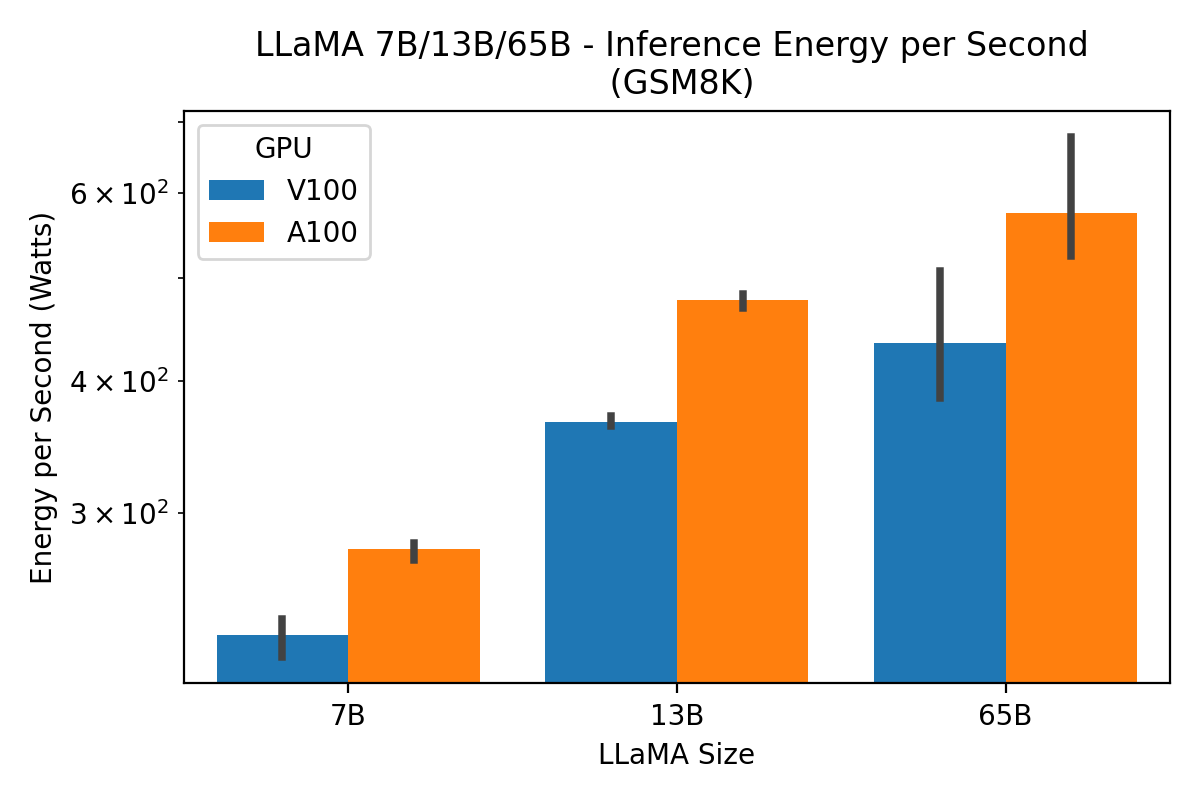} }}
    \caption{\textbf{Baseline energy per second (Watts) estimates of performing inference with LLaMA 7B, 13B, and 65B}: inference energy comparisons on the minimum set of hardware/settings required (see Table \ref{tab:limits}) with Alpaca and GSM8K on a log-scale. Color indicates device (V100/A100), bars indicate average quantities and lines indicate error bars. Energy is averaged over maximum generation lengths of 256, 512, and 1024 due to near-identical energy/size trends for each generation length.}
    \label{fig:energy-baseline}
\end{figure}

%Turning to the energy costs of inference for LLaMA 7B, 13B, and 65B, 
Figure \ref{fig:energy-baseline} shows a comparison of the energy per second required to run inference on LLaMA 7B, 13B, and 65B,  with different GPUs under the same bare minimum hardware settings as the above. For both the Alpaca and GSM8K datasets, we see that there is a considerable increase in the energy per second across all LLaMA sizes when using the A100 over the V100 where the most considerable increase is for the smallest 7B model. Although Figure \ref{fig:baselines} shows a considerable increase in inference throughput from using the A100, Figure \ref{fig:energy-baseline} shows us that this improvement does not come for free: it comes at an increased energy cost per second. Moreover, for the largest LLaMA 65B, it is less clear whether the increased inference energy per second (Figure \ref{fig:energy-baseline}) is worth the small improvement in inference throughput in terms of words/token/responses per second (Figure \ref{fig:baselines}).

%This figure also shows the energy consumed when we increase the maximum generated response length from 256 to 512 to 1024. As expected, we see an increase in the energy consumed as the model output length is increased. For the smaller LLaMA models (7B, 13B), the computational efficiencies and increased performance of A100 GPU mean that the total inference runs significantly faster and results in much lower energy consumption compared with the V100 GPU. On the other hand, in the case of the 65B parameter model, the results are more interesting. Running the 65B model on the V100 GPUs requires double the number of GPUs compared with the A100 GPU but this does not result in a doubling of the energy energy used. We observe only a marginal increase in the energy used for the same workload. Finally, across all models, increasing the maximum response generation length beyond 512 does not require an increasing amount of energy. This may be due to the fact  that the responses for the queries selected for this experiment did not require the model to generate longer responses and is an area of further investigation. For instance, changing model parameters such as the temperature and top-$p$ could potentially affect these results. 

\subsection{Energy per Second: LLaMA 65B}
\begin{figure}
    \centering
    {{\includegraphics[width=7cm]{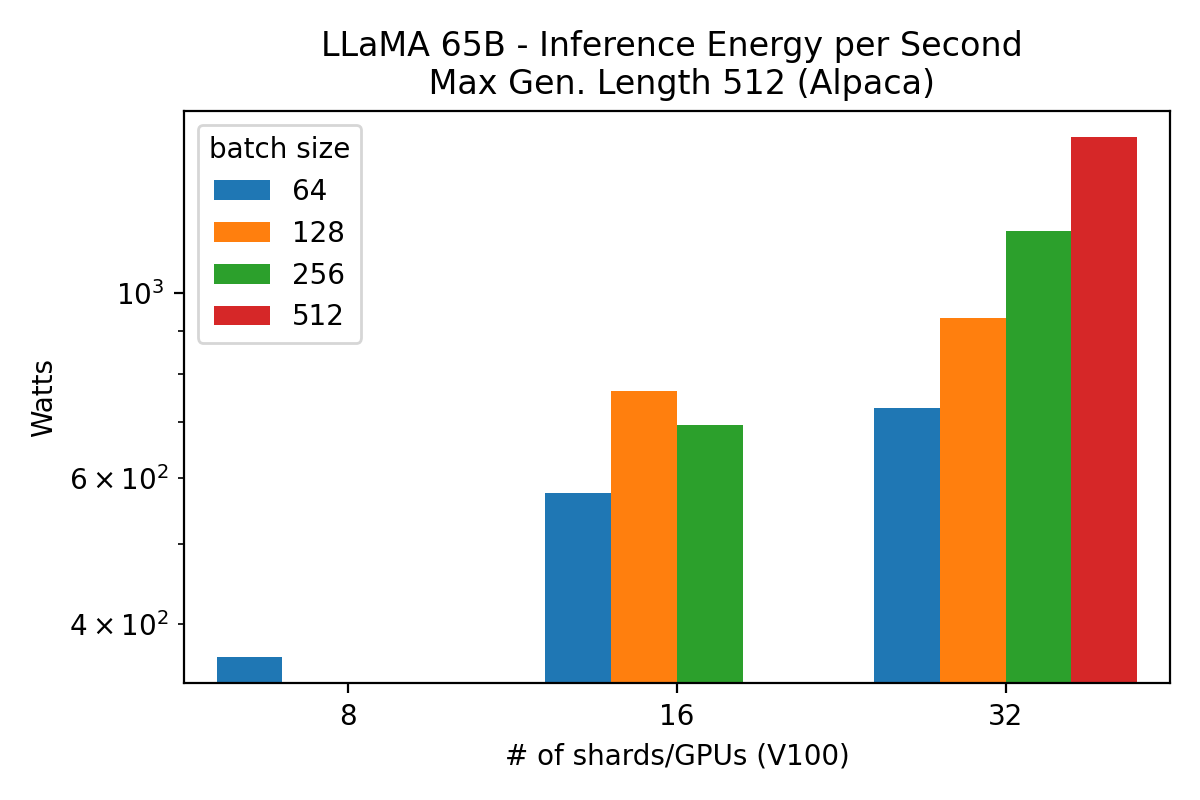} }}
    \qquad
    {{\includegraphics[width=7cm]{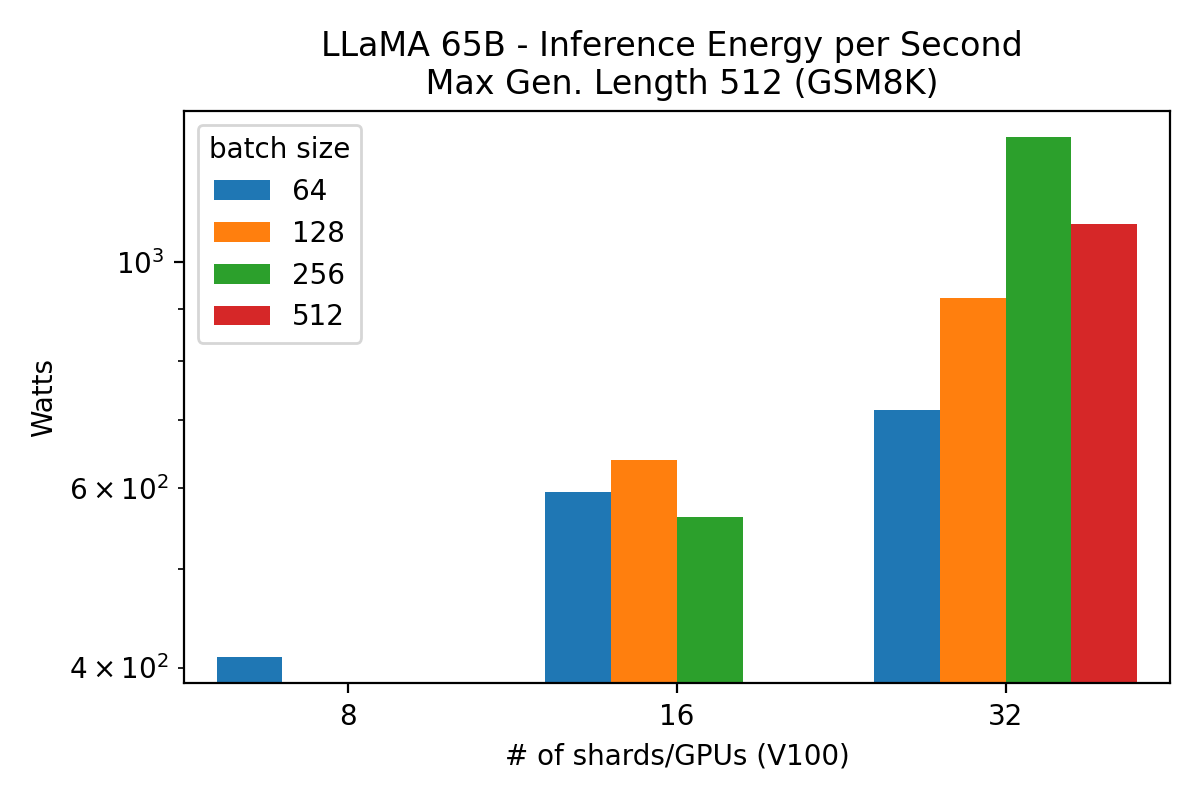} }}
    \caption{\textbf{Energy per second (Watts) estimates of LLaMA 65B across batch sizes of 64/128/256/256 and 8/16/32 shards for max generation length 512}: inference energy estimates on Alpaca and GSM8K on log-scale. Color indicates batch size.}
    \label{fig:65b_energysecond_gen512}
\end{figure}

\begin{figure}[htbp!]
    \centering
    {{\includegraphics[width=7cm]{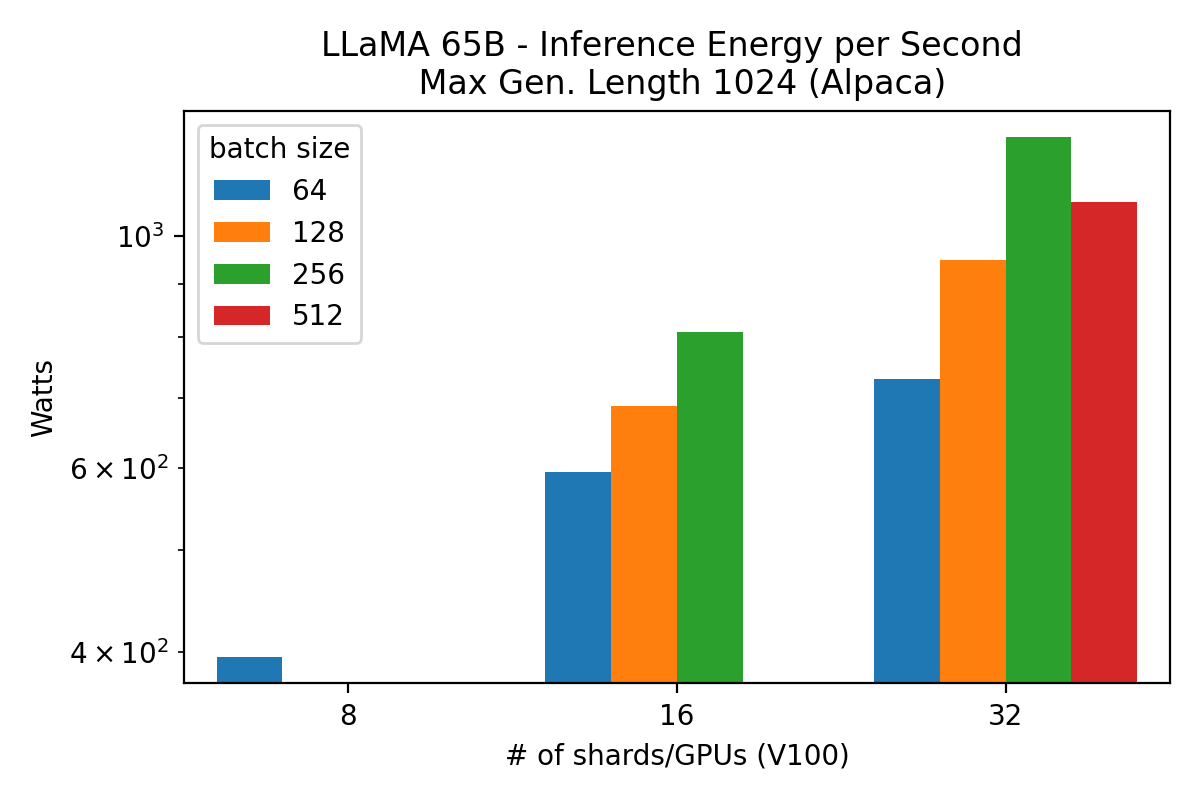} }}
    \qquad
    {{\includegraphics[width=7cm]{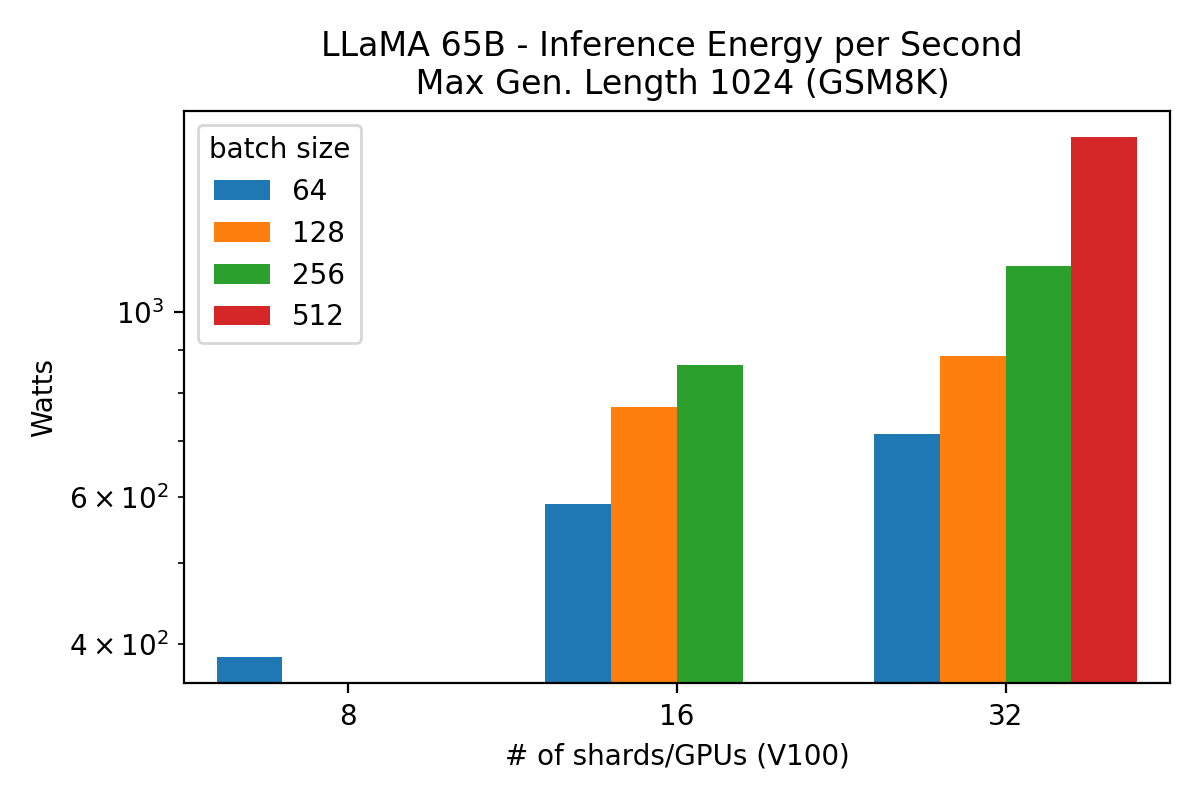} }}
    \caption{\textbf{Energy per second (Watts) estimates of LLaMA 65B across batch sizes of 64/128/256/512 and 8/16/32 shards for max generation length 1024}: inference energy estimates on Alpaca and GSM8K on log-scale. Color indicates batch size.}
    \label{fig:65b_energysecond_gen1024}
\end{figure}

We first take a look at the amount of energy inference costs per unit time in seconds. Figures \ref{fig:65b_energysecond_gen512} and \ref{fig:65b_energysecond_gen1024} show a more in-depth look of the energy inference costs of LLaMA 65B across different batch sizes and degrees of sharding. Specifically, Figure \ref{fig:65b_energysecond_gen512} shows energy costs for maximum generation length 512 and Figure \ref{fig:65b_energysecond_gen1024} shows energy costs for 1024.

Overall, we see an average increase in energy per second with the number of shards. While there is a slight correlation as energy per second increases with increasing batch size, increasing the number of shards always increases the wattage. Indeed, the energy per second increases with the number of shards even at the same batch size (e.g., the energy of inference at batch size 64, going from 16 shards to 32 shards). For both datasets, increasing the max generation length from 512 to 1024 does seem to increase the energy per second for each batch size within each shard configuration, but the overall effect is less clear or consistent. Overall, we see that the energy per second for inference with LLaMA 65B is on the order of 300 Watts to 1 Kilowatt from the lower shard configuration of 8 GPUs to the higher end of 32 GPUs.

\subsection{Energy per Decoded Token: LLaMA 65B}

\begin{figure}
    \centering
    {{\includegraphics[width=7cm]{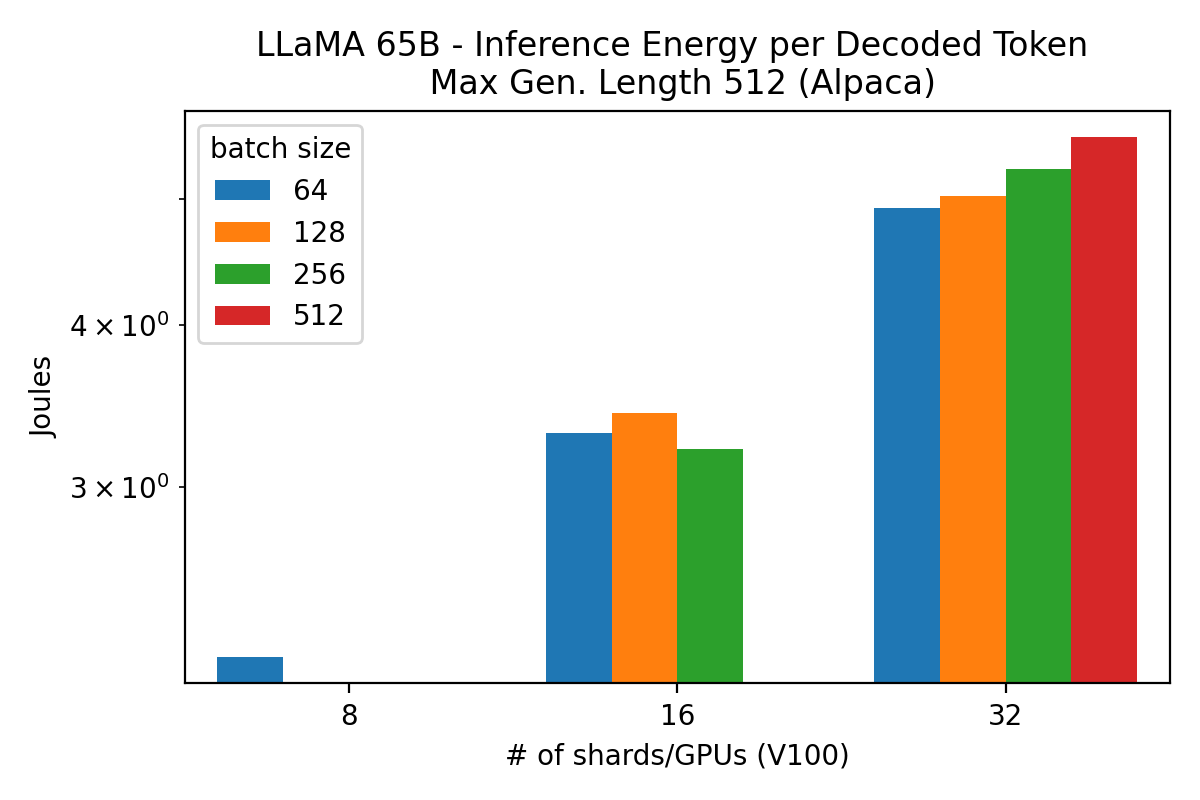} }}
    \qquad
    {{\includegraphics[width=7cm]{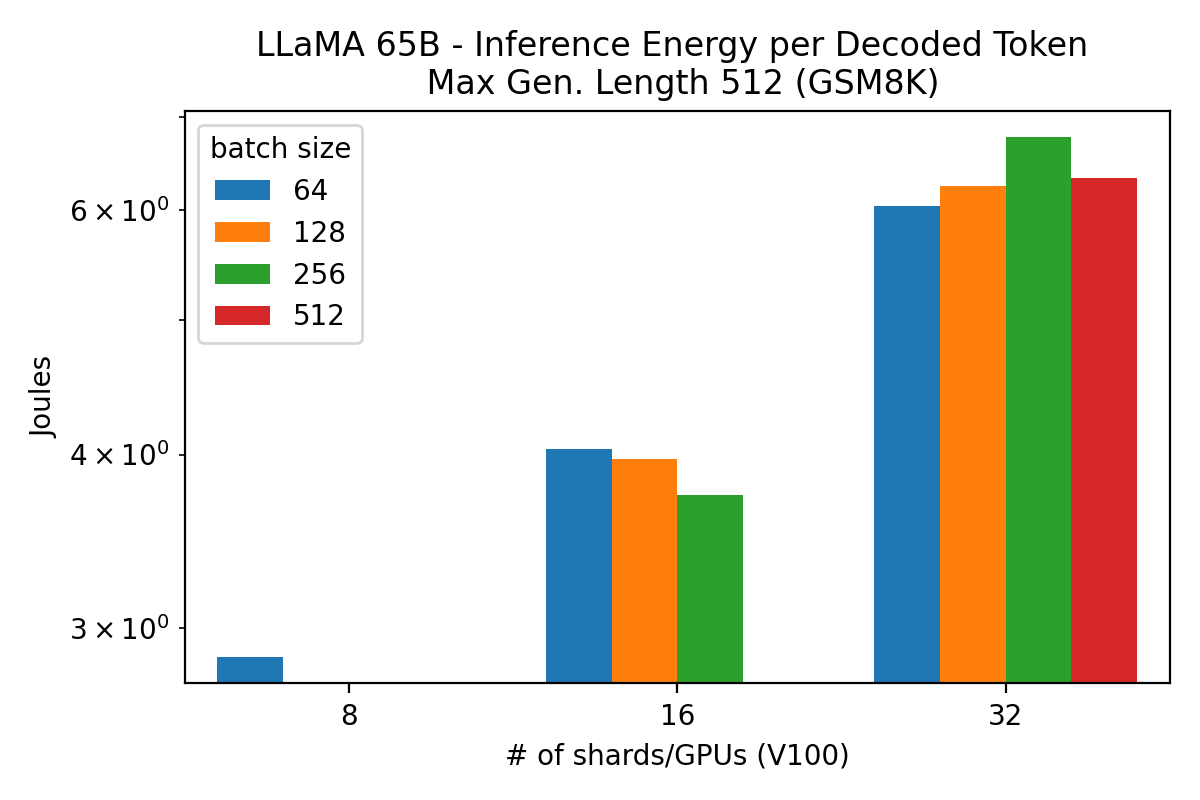} }}
    \caption{\textbf{Energy per output token estimates of LLaMA 65B across batch sizes of 64/128/256/512 and 8/16/32 shards for max generation length 512}: inference energy estimates on Alpaca and GSM8K on log-scale. Color indicates batch size.}
    \label{fig:65b_energytoken_gen512}
\end{figure}

\begin{figure}
    \centering
    {{\includegraphics[width=7cm]{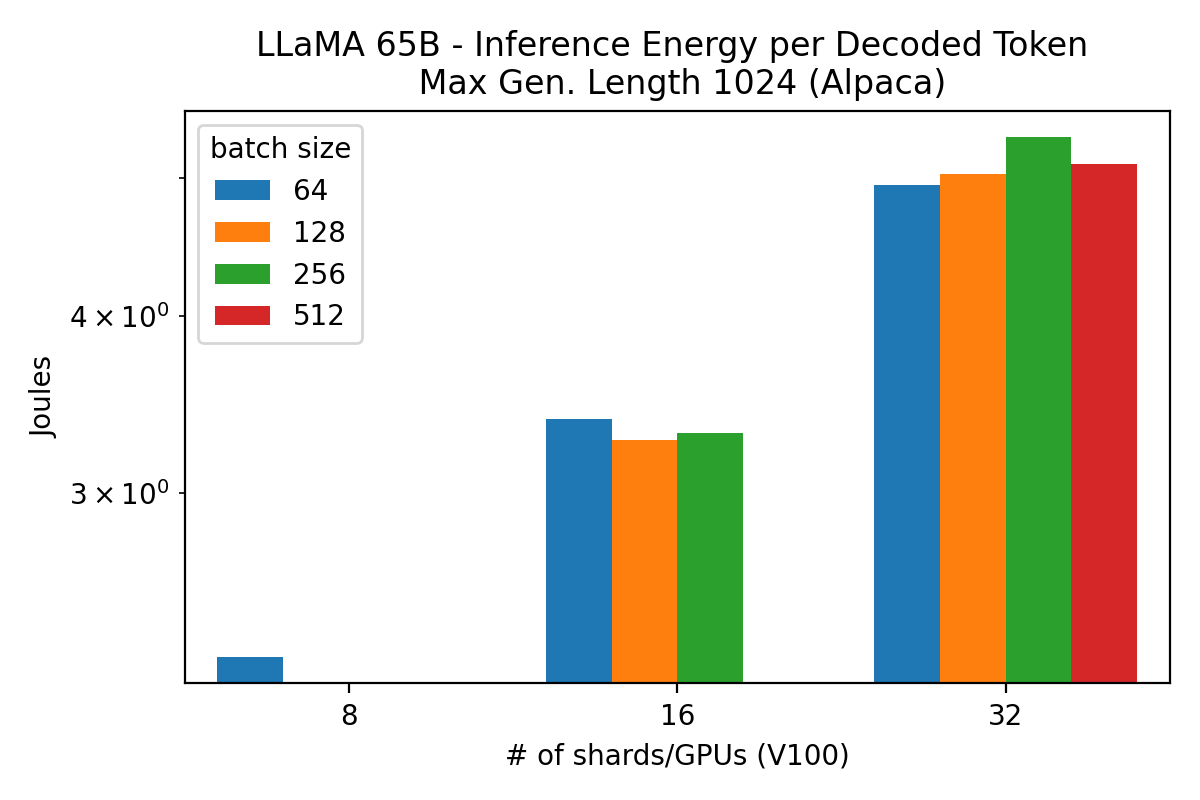} }}
    \qquad
    {{\includegraphics[width=7cm]{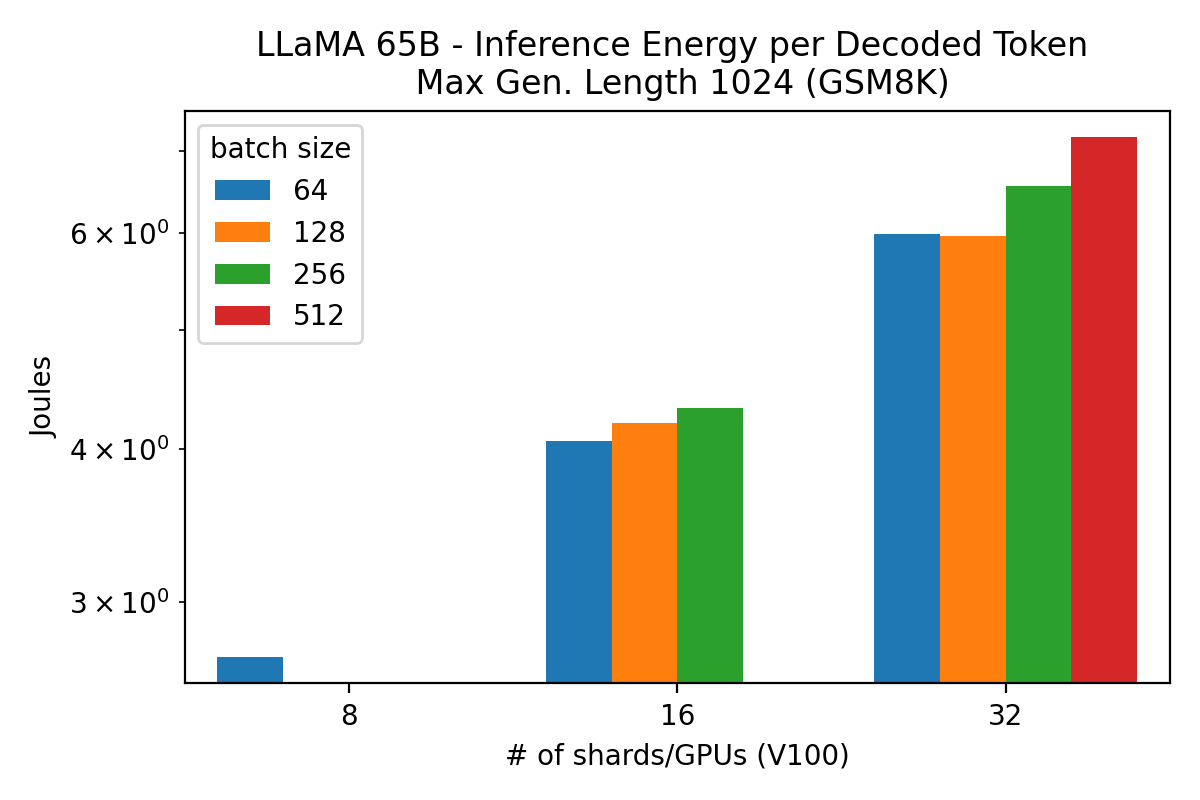} }}
    \caption{\textbf{Energy per output token estimates of LLaMA 65B across batch sizes of 64/128/256/512 and 8/16/32 shards for max generation length 1024}: inference energy estimates on Alpaca and GSM8K on log-scale. Color indicates batch size.}
    \label{fig:65b_energytoken_gen1024}
\end{figure}

Moving on to energy per each decoded output token, we see that in Figures \ref{fig:65b_energytoken_gen512} and \ref{fig:65b_energytoken_gen1024} that energy per token tends to follow a similar pattern in relation to the number of shards: as the number of shards increases, the energy per output token also increases. However, we see little change in the average energy per token between max generation length 512 and 1024. For instance, with length 512, we see that it takes about 3-4 Joules for a output token, which is approximately the same amount for length 512. As with energy per second, max generation length seems to have a negligible effect on energy costs from 512 to 1024. Interestingly, there appears to be an exception for the GSM8K math problem dataset; there exists a ``sweet spot'' at 16 shards where continuously increasing the batch size can actually reduce the energy per token at max generation length 512. However, this disappears under max generation length 1024 where increasing the batch size increases the energy per token. The definitive existence of this sweet spot for datasets of differing styles/complexities, or others like it, will require more experimentation and benchmarking to establish.

\subsection{Energy per Response: LLaMA 65B}
\begin{figure}
    \centering
    {{\includegraphics[width=7cm]{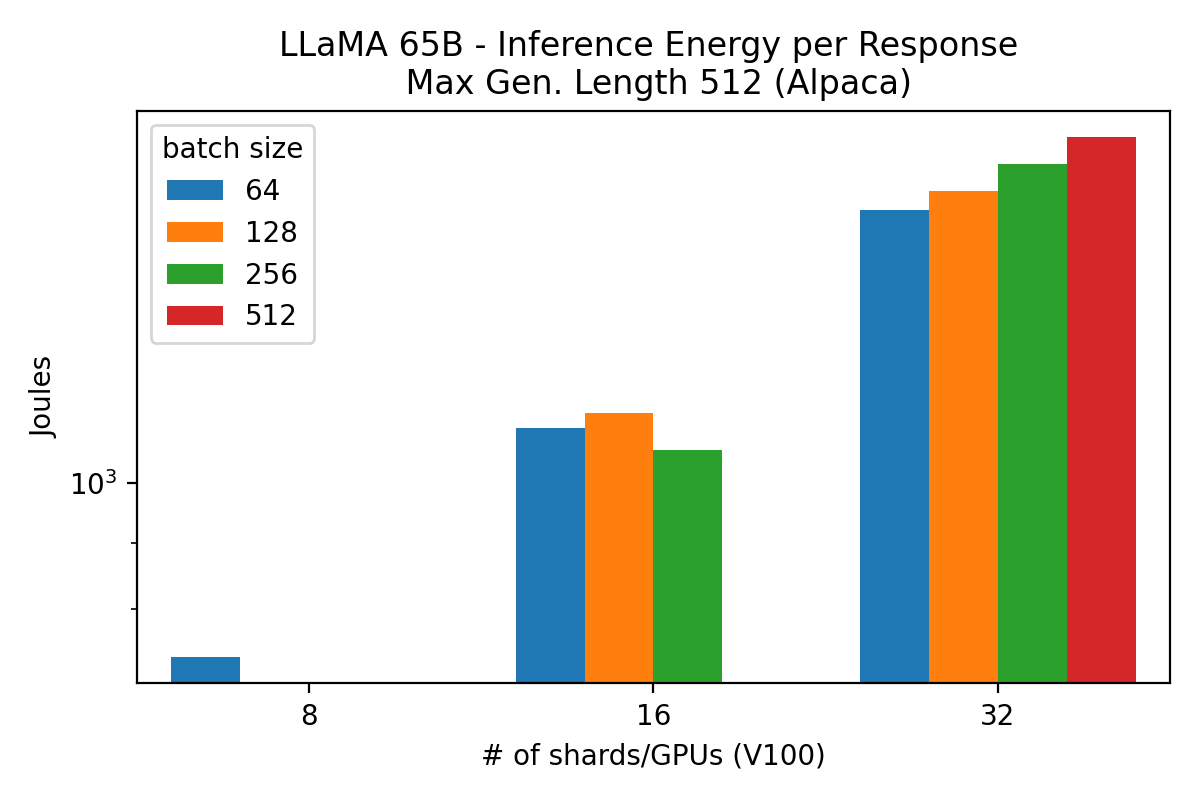} }}
    \qquad
    {{\includegraphics[width=7cm]{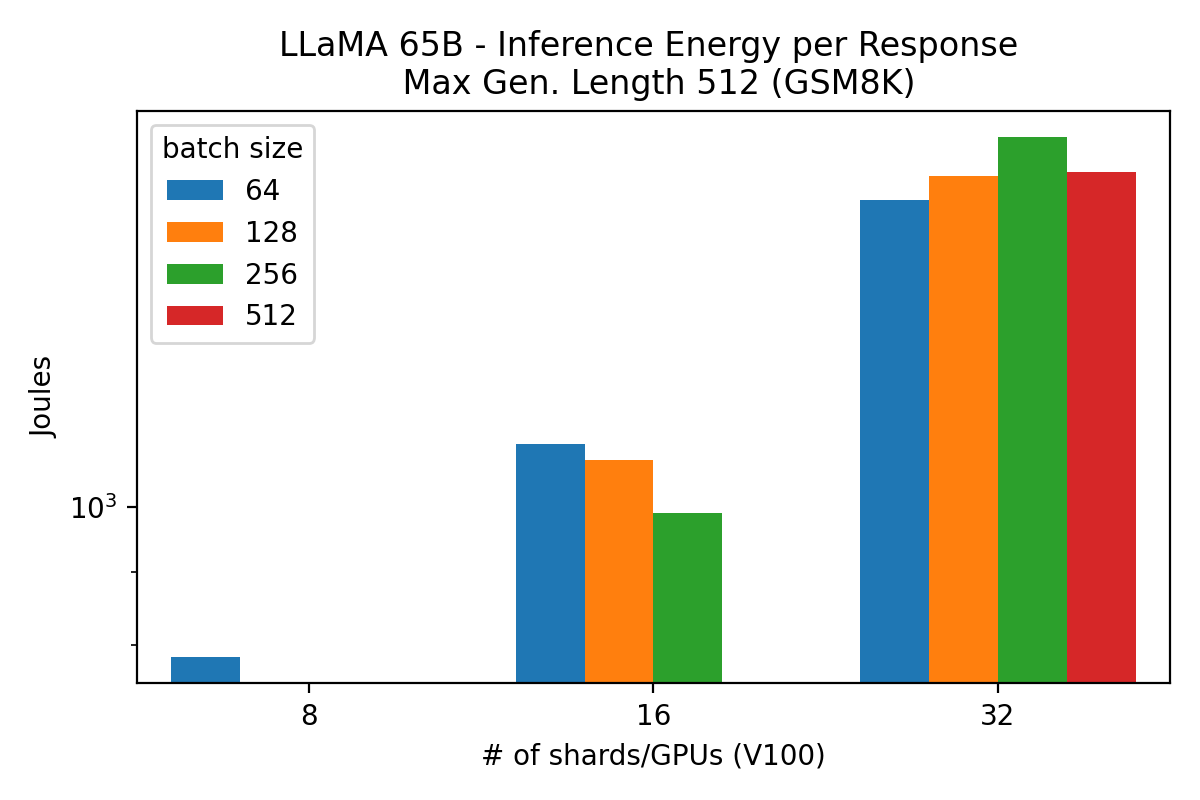} }}
    \caption{\textbf{Energy per response estimates of LLaMA 65B across batch sizes of 64/128/256/512 and 8/16/32 shards for max generation length 512}: inference energy estimates on Alpaca and GSM8K on log-scale. Color indicates batch size.}
    \label{fig:65b_energyresponse_gen512}
\end{figure}

\begin{figure}
    \centering
    {{\includegraphics[width=7cm]{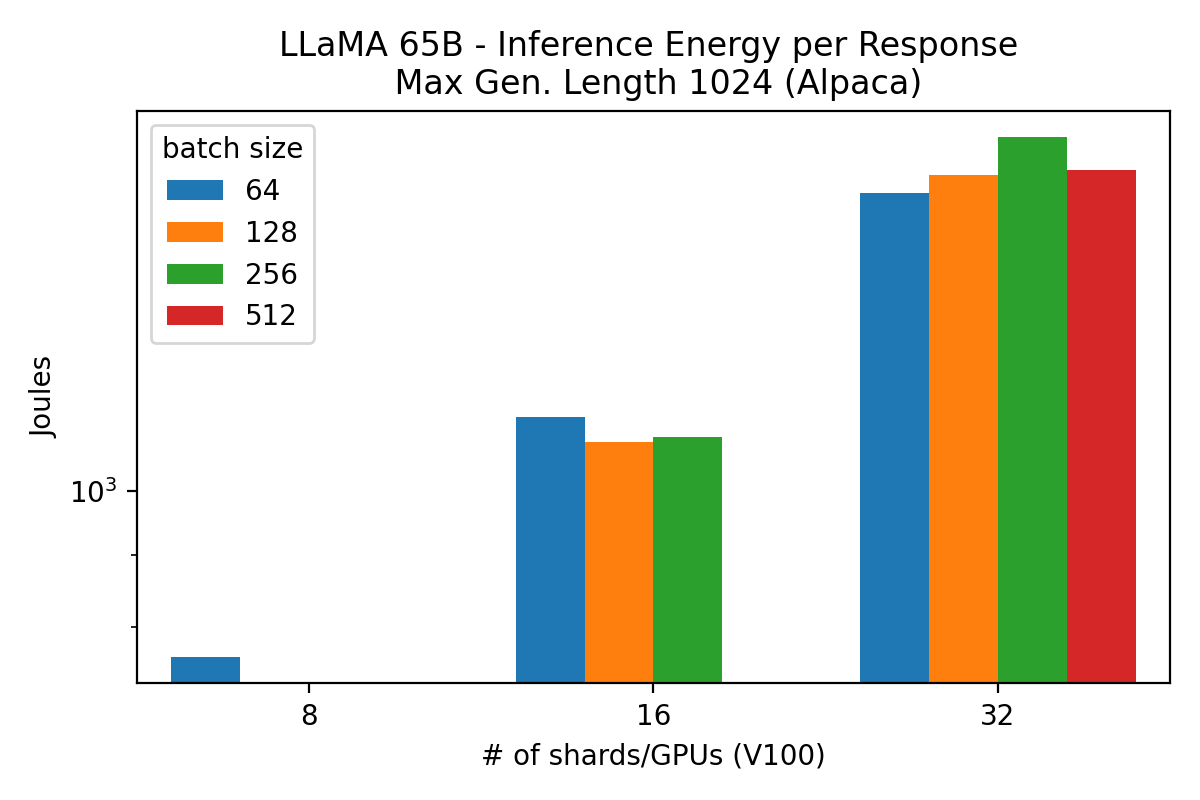} }}
    \qquad
    {{\includegraphics[width=7.4cm]{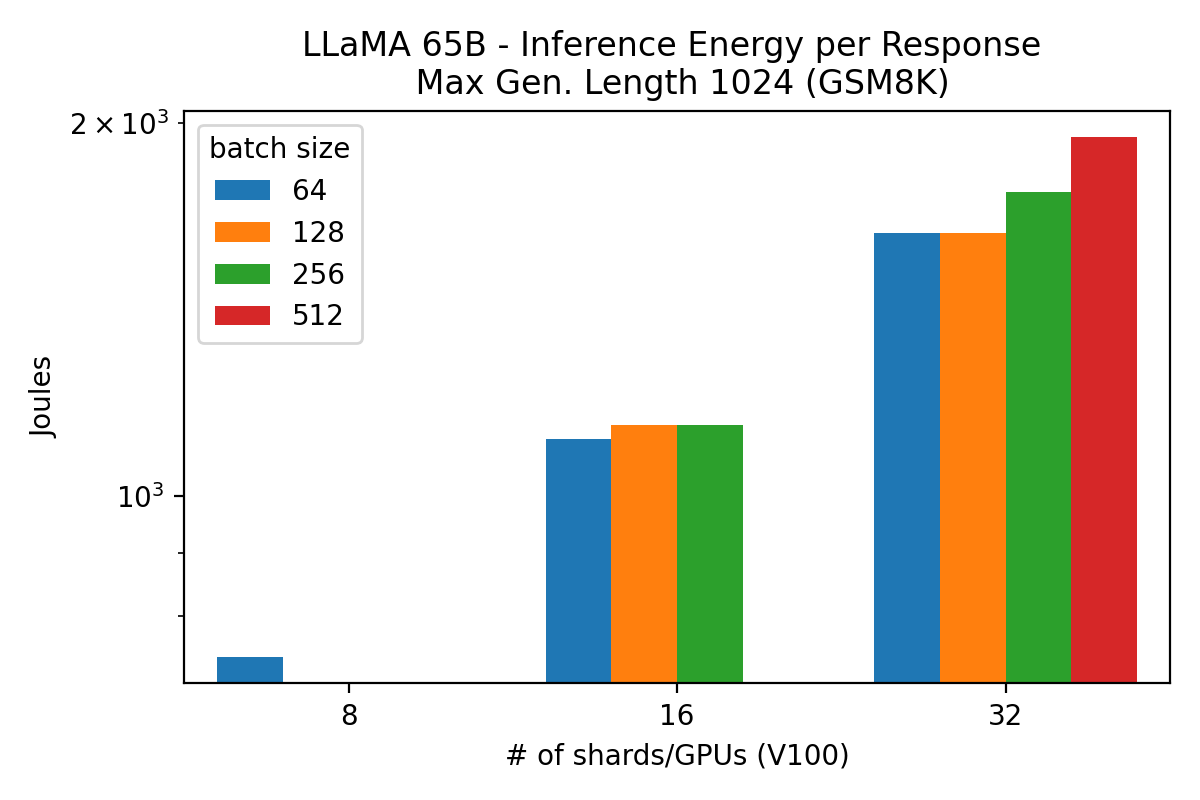} }}
    \caption{\textbf{Energy per response  estimates of LLaMA 65B across batch sizes of 64/128/256/512 and 8/16/32 shards for max generation length 512}: inference energy estimates on Alpaca and GSM8K on log-scale. Color indicates batch size.}
    \label{fig:65b_energyresponse_gen1024}
\end{figure}

Figures \ref{fig:65b_energyresponse_gen512} and \ref{fig:65b_energyresponse_gen1024} show energy metrics in terms of responses from the 65B model. Like before, we see that increasing the number of shards still tends to increase the energy costs of inference per response most overall while increasing the maximum generation length from 512 (Figure \ref{fig:65b_energyresponse_gen512}) to 1024 (Figure \ref{fig:65b_energyresponse_gen1024}) does not induce a clear or significant effect in inference energy costs. 
Also like before, while we see slight increases in energy costs per response generated within a shard configuration as batch size increases, but not consistently or significantly. Again, we see that for GSM8K, at max generation length 512, increasing the batch size while keeping the number of shards fixed at 16 is associated with a decrease in energy per response, which is consistent with what we observed in energy per tokens in the same setting.

\subsection{Effects of GPU Power Capping on LLaMA 65B}
Power consumption in AI is an increasingly important concern. In prior work, we have shown~\cite{mcdonald2022} that power capping GPUs during training of language models such as BERT~\cite{bert} is an effective way of reducing the energy consumed training these models. While the work in ~\cite{mcdonald2022} focused on model training, in this paper, we focus on inference. In order to study the effect of power capping on inference using large language models, we ran a limited set of experiments using LLaMA 65B. We ran the 65B model on four 80GB A100 GPUs with the power cap set at 250W, 175W and 150W. 

Table~\ref{tab:power-cap} shows the relative change in total inference time, energy and token rate under power cap conditions. Results shown here are calculated relative to a power cap of 250W. For a 30\% reduction in power from 250W to 175W, the inference time increases by an average of 6.7\% for a corresponding average reduction in total energy by 23.21\%. However, a reduction in power cap to 150W results in a much more significant (19.49\%) increase in average inference time. These results show that power capping as an energy savings intervention can be effective when applied appropriately. A static power cap for all GPU workloads may not show the same effectiveness depending on the task and additional experimentation is required to make broader recommendations. 

\begin{table}[h]
    \centering
    \begin{tabular}{lcccccccc}
    \toprule
        Output & \multicolumn{2}{c}{Time} && \multicolumn{2}{c}{Energy} && \multicolumn{2}{c}{Token Rate}\\
        length & \multicolumn{2}{c}{\% change} && \multicolumn{2}{c}{\% change} && \multicolumn{2}{c}{\% change} \\
    \midrule
                             & 175W & 150W & & 175W & 150W & & 175W & 150W \\ 
         \cline{2-3} \cline{5-6} \cline{8-9} \\
         256                & 6.23 & 15.33  & & -21.82 & -32.76 & & -5.87 & -13.15 \\
         512                & 6.51 & 21.70 & & -23.95 & -34.66 & & -6.11 & -17.83 \\
         1024               & 7.40 & 21.65 & & -23.87 & -34.59 & & -6.89 & -17.80 \\
    \bottomrule
    \end{tabular}
    \caption{\textbf{Effects of GPU power capping on LLaMA 65B inference}: This table shows the relative performance of the LLaMA 65B model on the GSM8k dataset with a batch size of 64 and output lengths of 256, 512, 1024 using NVIDIA A100 GPUs. The GPUs were power capped at 250W, 175W and 150W. Results shown here are relative to model performance at 250W to stay consistent with the settings in the rest of the experiments described here.}% As shown in the table, a 30\% reduction in GPU power reduces model performance by only 6\% but leads to nearly a 23\% reduction in energy consumed for the same workload. This simple intervention has the potential to help reduce datacenter GPU energy consumption with only a small reduction incompute performance for appropriate workloads.}
    \label{tab:power-cap}
\end{table}

\subsection{GPU Resource Utilization under Distributed Inference}
Finally, we briefly examine the average GPU resource utilization by the 65B LLaMA model when running model sharded inference. For the sake of simplicity, we only consider a batch size of 64 and a maximum generated output length of 256. For this configuration, we ran on four A100 GPUs and 8, 16, 32 V100 GPUs. These results are summarized in Tables~\ref{tab:gpu_util} and ~\ref{tab:v100_util}. In all cases, the streaming multiprocessors (SM) utilization as reported by the \texttt{DCGM} utility was observed to be in the 94\%-95\% range. For the A100 GPUs, the average SM utilization rises to 98\% when the maximum generated output length is increased to 2048. Given that the model is sharded in a manner that enables us to load it fully in GPU memory and run inference on a non-trivial amount of data, we expect memory utilization to be low depending on the specific model parameters and input sizes used. Thus, on the four 80GB A100 nodes, the memory utilization varies between 23\%-27\% depending the maximum generated output length. This under-utilization of memory implies that it may be possible to co-locate multiple models on the same set of GPUs to increase aggregate throughput and potentially reduce cloud compute costs or improve system utilization at a supercomputer center. With new GPU sharing capabilities such as Multi-Process Service (MPS)~\cite{mps} and Multi-Instance GPU (MIG)~\cite{mig}, a single GPU may be shared by diverse workloads for an overall improvement in system throughput as shown in recent work~\cite{li2022miso}. The optimal GPU configuration for sharing LLMs and other workloads is a part of our future work in this area. 

\begin{table}[h]
    \centering
    \begin{tabular}{cccc}
    \toprule
        Model Shards & Output Length & Max. Memory  Util. &  Avg. SM Util. \\
                      \midrule
          4 & 256 &     23.36 &   95.00 \\
          4 & 512  &   24.54  &  98.81 \\
         4 & 1024   &  24.85  &  98.85 \\
         4 & 2048    & 27.00  &  98.00 \\
    \bottomrule
    \end{tabular}
    \caption{\textbf{A100 Utilization:} This table shows GPU utilization for 80GB A100 GPUs and LLaMA 65B with 4 shards, batch size of 64 averaged across both datasets used in this paper.}
    \label{tab:gpu_util}
\end{table}

\begin{table}[h]
    \centering
    \begin{tabular}{cccc}
    \toprule 
    Model Shards &  Output Length & Max. Memory Util.  & Avg. SM Util.  \\
    \midrule
         8    & 256 & 24.25   & 94.75    \\      
          16  & 256    & 13.33  &  95.00  \\       
          32   & 256   &  6.66   & 95.66 \\
    \bottomrule
    \end{tabular}
    \caption{\textbf{V100 Utilization:} This table shows GPU utilization for 32GB V100 GPUs and LLaMA 65B with 8, 16, 32 shards, a batch size of 64 and maximum generated output length of 256 averaged across both datasets used in this paper. We limit this result to an ouptut length of 256 because longer outputs on 8 V100 GPUs are not possible given memory limits of the GPU.}
    \label{tab:v100_util}
\end{table}
% \begin{figure*}[ht]
%     \centering
%     \includegraphics[width=0.98\linewidth]{figures/v100_65b_comparison_timing_bar_gsm.png}
%     \caption{LLaMA 65B performance comparison on V100 GPUs: This figure shows a comparison of the LLaMA 65B model on 8, 16 and 32 GPUs. Each model used a batch size of 64 and the maximum output length tested was 256, 512 and 1024. As the number of GPUs used increases, the model performance reduces because the communication costs start becoming more significant.\sid{use this one or Figure~\ref{fig:65b-v100-only}. this one might be too large and unwieldy}}
%     \label{fig:65b-v100-only-bar}
% \end{figure*}

% \sid{
% \begin{enumerate}
%     %\item compare effect of power capping on 8 shards on 80GB A100 on both datasets
%     %\item plot some time series data from GPUs to illustrate under-utilization
%     \item some thoughts on how we could use workload co-location to improve GPU utilization/throughput
% \end{enumerate}
% }

\section{Discussion}
\label{sec:discussion}
In this paper, we show the results of benchmarking a representative large language model on NVIDIA GPUs. We show baseline results from smaller models (7B, 13B) and compare them against the largest available version (65B) of LLaMA. We also examine the inference performance and energy across distributed settings and different configurations by varying model parameters, input data, and hardware configurations. By comparing a natural language instruction following dataset (Alpaca) and a mathematical question-answer dataset (GSM8K), we also find that the complexity of the input dataset can affect the model performance for a given set of hyper-parameters and hardware configuration. %This hints to the added complexity of optimizing language models across different applications. 

Given the size of LLMs and the limits imposed by current hardware, inference with large models can impose onerous requirements. For example, we find that, at a minimum, 8 V100 GPUs each with 32 GB of RAM or 4 A100 GPUs each with 80GB of memory are required for any meaningful inferences with the 65B LLaMA model. In each case among our experiments, we shard the model evenly across all GPUs in order to fit the model/data; however, this results in only 20\%-25\% of the GPU memory being utilized at any given time. This over-provisioning of resources represents new opportunities for resource sharing across multiple workloads in the latest NVIDIA GPUs. The Multi-Process Service (MPS)~\cite{mps} and Multi-Instance GPU (MIG)~\cite{mig} are new capabilities that enable GPU sharing across different workloads. Although identifying the optimal MPS or MIG configuration for a given set of workloads is challenging, recent work~\cite{li2022miso} has developed new techniques to exploit these capabilities in order to dynamically partition GPU resources. This opens up the potential to optimally partition high-end GPUs such as the A100s or H100s to co-locate multiple LLMs for inference---with the potential of only minimal degradation to computational performance. 

Finally, as AI compute requirements have increased, there is an increasing focus on approaches to reduce the carbon and energy footprints of datacenters by making larger models leaner or more efficient. Approaches such as model quantization, distillation, sparsification, etc. are being developed to reduce the compute required for AI along with the development of custom, energy-efficient hardware for inference and training. However, simple interventions like GPU power capping is available to be deployed today---our preliminary analysis with LLM inference in this paper suggests that power capping can be an effective tool for reducing inference energy. If applied at the datacenter-scale, this intervention has the potential to reduce overall energy usage in the long-run as new approaches are developed to address the energy consumption of AI compute.  

As part of our future plans, we aim to conduct similar experiments on other open-source, large language models along with more in-depth characterization of compute and energy for not just inference, but also for the training/fine-tuning of these models. It is our hope that this paper provides a baseline for inference with LLMs and fosters a broader discussion of the challenges and opportunities in this field.

\section*{Acknowledgements}
The authors acknowledge the MIT SuperCloud \cite{reuther2018interactive} and Lincoln Laboratory Supercomputing Center for providing HPC and consultation resources that have contributed to the research results reported within this paper. The authors acknowledge the MIT SuperCloud team: William Arcand,  William Bergeron, Chansup Byun,  Michael Houle, Anna Klein, Peter Michaleas, Lauren Milechin, Julie Mullen, Albert Reuther, Antonio Rosa, and Charles Yee. The authors also wish to acknowledge the following individuals for their contributions and support: Bob Bond, Allan Vanterpool, Tucker Hamilton, Jeff Gottschalk, Mike Kanaan, Charles Leiserson, Dave Martinez, Steve Rejto, Marc Zissman.

\balance
%\newpage
\bibliographystyle{IEEEtran} 
\bibliography{IEEEabrv,references.bib}
\end{document}